\documentclass[10pt,twocolumn,letterpaper]{article}

\usepackage{iccv}
\usepackage{times}
\usepackage{epsfig}
\usepackage{graphicx}
\usepackage{amsmath}
\usepackage{amssymb}

\usepackage{import}
\usepackage{algorithm}
\usepackage{algpseudocode}
\usepackage{float}
\usepackage[font=small]{caption}
\usepackage[dvipsnames]{xcolor}
\usepackage[normalem]{ulem}

\usepackage[breaklinks=true,bookmarks=false,colorlinks]{hyperref}

\usepackage{booktabs}
\usepackage{array}
\newcolumntype{L}[1]{>{\raggedright\let\newline\\\arraybackslash\hspace{0pt}}m{#1}}
\newcolumntype{C}[1]{>{\centering\let\newline\\\arraybackslash\hspace{0pt}}m{#1}}
\newcolumntype{R}[1]{>{\raggedleft\let\newline\\\arraybackslash\hspace{0pt}}m{#1}}
\usepackage{tablefootnote}
\usepackage[font=small]{caption}
\usepackage{subcaption}
\usepackage{multirow}
\usepackage{enumitem}
\usepackage{float}
\usepackage{xcolor}
\usepackage{array}

\definecolor{pros}{HTML}{00A000}
\definecolor{cons}{HTML}{FF0000}
\definecolor{neutral}{HTML}{FFCC00}

\iccvfinalcopy 

\def\iccvPaperID{****} 

\begin{document}


\title{Forget-Me-Not: Learning to Forget in Text-to-Image Diffusion Models}

\author{
    Eric Zhang\textsuperscript{1*}, 
    Kai Wang\textsuperscript{1*}, 
    Xingqian Xu\textsuperscript{1,3}, 
    Zhangyang Wang\textsuperscript{2,3}, 
    Humphrey Shi\textsuperscript{1,3} \\
    {\small \textsuperscript{1}SHI Labs @ U of Oregon \& UIUC, \textsuperscript{2}UT Austin, \textsuperscript{3}Picsart AI Research (PAIR)}\\
    {\small \textbf{\url{https://github.com/SHI-Labs/Forget-Me-Not}}}
}

\twocolumn[{
\maketitle
\begin{center}
    \captionsetup{type=figure}
    \includegraphics[width=0.99\textwidth]{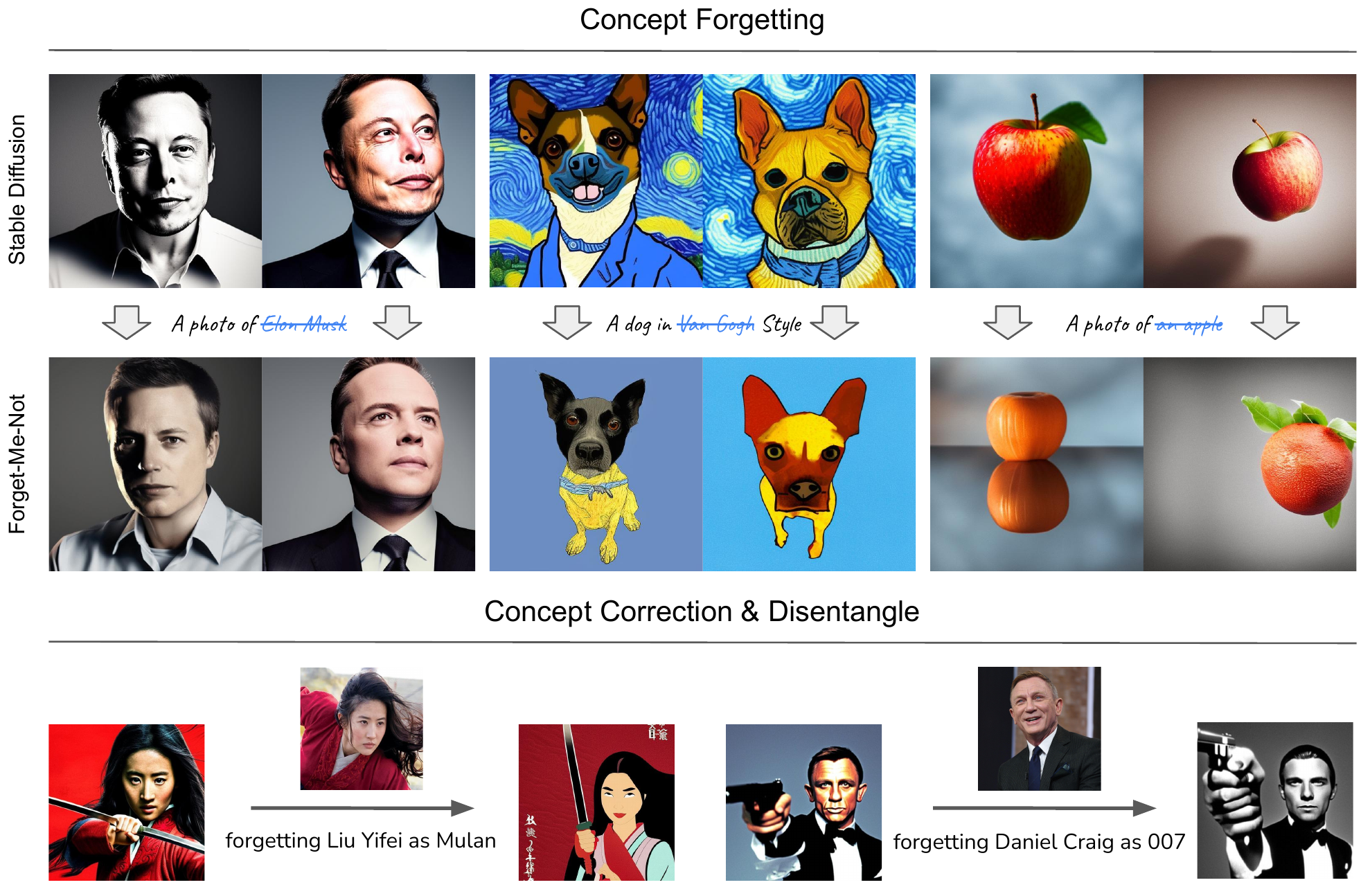} 
    \captionof{figure}{Given a text-to-image model (\ie Stable Diffusion), our approach can swiftly re-steer the cross attention towards a specific concept and subsequently forgetting or correcting it.
    (1) Concept Forgetting: target concepts (denoted in blue text and crossed-out) are successfully removed without compromising the quality of the output. (2) Concept Correction \& Disentangle: our method can be used to correct a dominant or undesired concept of a prompt. Prior overshadowed concepts reveal in outputs after the dominant concepts are forgotten. In addition, our method learns to forget fast with only 30 seconds for certain concepts (e.g. Elon Musk), and can be easily adapted to lightweight model patches for Stable Diffusion, allowing for multi-concept manipulation and convenient distribution to users.
    }
    \label{fig:teasor}
\end{center}
}]

\ificcvfinal\thispagestyle{empty}\fi

\def\thefootnote{*}\footnotetext{Equal contribution}\def\thefootnote{\arabic{footnote}}


\begin{abstract}

The unlearning problem of deep learning models, once primarily an academic concern, has become a prevalent issue in the industry. The significant advances in text-to-image generation techniques have prompted global discussions on privacy, copyright, and safety, as numerous unauthorized personal IDs, content, artistic creations, and potentially harmful materials have been learned by these models and later utilized to generate and distribute uncontrolled content. To address this challenge, we propose \textbf{Forget-Me-Not}, an efficient and low-cost solution designed to safely remove specified IDs, objects, or styles from a well-configured text-to-image model in as little as 30 seconds, without impairing its ability to generate other content. Alongside our method, we introduce the \textbf{Memorization Score (M-Score)} and \textbf{ConceptBench} to measure the models' capacity to generate general concepts, grouped into three primary categories: ID, object, and style. Using M-Score and ConceptBench, we demonstrate that Forget-Me-Not can effectively eliminate targeted concepts while maintaining the model's performance on other concepts. 
Furthermore, Forget-Me-Not offers two practical extensions: a) removal of potentially harmful or NSFW content, and b) enhancement of model accuracy, inclusion and diversity through \textbf{concept correction and disentanglement}. It can also be adapted as a lightweight model patch for Stable Diffusion, allowing for concept manipulation and convenient distribution. To encourage future research in this critical area and promote the development of safe and inclusive generative models, we will open-source our code and ConceptBench at \href{https://github.com/SHI-Labs/Forget-Me-Not}{https://github.com/SHI-Labs/Forget-Me-Not}.

\end{abstract}


\section{Introduction}
Recently, text-to-image models~\cite{chang2023muse, make-a-scene, dalle1, dalle2, imagen, parti, Rombach_2022_CVPR, vd} have shown impressive performance in synthesizing high-quality images according to text prompts. Among these methods, diffusion models such as DALL-E 2~\cite{dalle2} and Stable Diffusion (SD)~\cite{Rombach_2022_CVPR} have met commercial-level productization requirements, initiating numerous applications for downstream users; such text-to-images are also recenlty shown to be able to generate and editing videos in a zero-shot fashion~\cite{text2videozero} without further training. Industrial solutions such as~\cite{picsart, adobefirefly, lexica, novelai, midjourny, diffusion-art-or-digital-forgery} have been widely adopted in various art and visual design systems, garnering significant public attention. 
Despite the popularity of this field, concerns about security, fairness, regulation, copyright, safety, etc., continue to grow rapidly in proportion to model usages. Risks such as generating unauthorized, biased, and unsafe content have become an immediate issue to be resolved. While this is not the first time the community has investigated these cases, prior efforts such as~\cite{fairdiffusion, fairnessGAN, A-Prompt-Array-Keeps-the-Bias-Away, Discover-and-Mitigate-Unknown-Biases} have proposed methods in which most of them are high-cost solutions focused on GAN. Yet we still need an effective and efficient solution that can be widely applied to large-scale diffusion models, which motivated us to dive deep into this topic.

The risks and issues associated with such large-scale text-to-image models originate from the billion-sized datasets used in training, including public datasets such as Laion~\cite{laion}, COYO~\cite{coyo}, CC12M~\cite{cc12m}, and private data from Google~\cite{imagen, parti}, OpenAI~\cite{dalle1, dalle2}, etc. The public datasets are usually web-scraped images and captions that lack human-level quality assurance on bias and safety, while private data sources are impossible to determine at scale. As a result, it is nearly unfeasible to fully address harmful content, privacy and copyright concerns through data filtering or source attribution. A compromised solution could be domain adaptation~\cite{Few-Shot-Diffusion-Models,Few-shot-Image-Generation-with-Diffusion-Models, Few-Shot-Generative-Model-Adaption}. In practice, people can adapt a large-scale model to a clean small/mid-size dataset and later use the model for image synthesis. However, collecting and filtering such datasets may still be quite laborious. Worse than that, such domain adaptation has severely influenced model capacity, making out-of-domain image synthesis challenging and sometimes nearly impossible. 

Will there be another path? Designing efficient methods and algorithms that guide existing large-scale text-to-image models to \textit{forget certain concepts} could be a better solution. 
We start this paper by first introducing this new mission, namely \textit{concept forgetting}, in which a designated set of concepts can be safely disentangled from the visual content. To achieve this goal, we proposed \textbf{Forget-Me-Not}, a simple, low-cost, but effective solution for concept forgetting. We also proposed the \textbf{memorization score (M-score)} along with \textbf{ConceptBench}, in which the former gauge the generative power of models on certain concepts, and the latter introduce sets of benchmark to assess concept forgetting and memorization. Moreover, we extend concept forgetting to \textit{concept correction \& disentangle} that may further assist models in being accurate and diverse. 

In conclusion, the main contribution of this paper can be summarized as the following: 
 
\begin{itemize}
    \item We propose \textbf{Forget-Me-Not}, a plug-and-play, efficient and effective concept forgetting and correction method for large-scale text-to-image models. 
    It provides an efficient way to forget specific concepts with as few as 35 optimization steps, which typically takes about 30 seconds. Additionally, Forget-Me-Not can be easily adapted as lightweight patches for Stable Diffusion, allowing for multi-concept manipulation and convenient distribution to text-to-image model users to address privacy, copyright, and safety concerns.

    \item We also propose \textbf{memorization score (M-score)} and \textbf{ConceptBench}, which enable quantitative measurements of models' capacity for synthesizing the target set of concepts. To the best of our knowledge, we are the first to introduce a numerical solution to gauge the model's behavior of memorization and forgetting. 

    \item Through extensive studies and tests, we demonstrate that our Forget-Me-Not is simple, low-cost, and effective. Downstream applications, such as harmful and NSFW content removal and biased concept correction \& disentangle, further expand our scope beyond concept forgetting towards cross-modal model refinements that may better fit real-world use cases.
\end{itemize}

\begin{figure*}[htbp]
    \centering
    \includegraphics[width=1\textwidth]{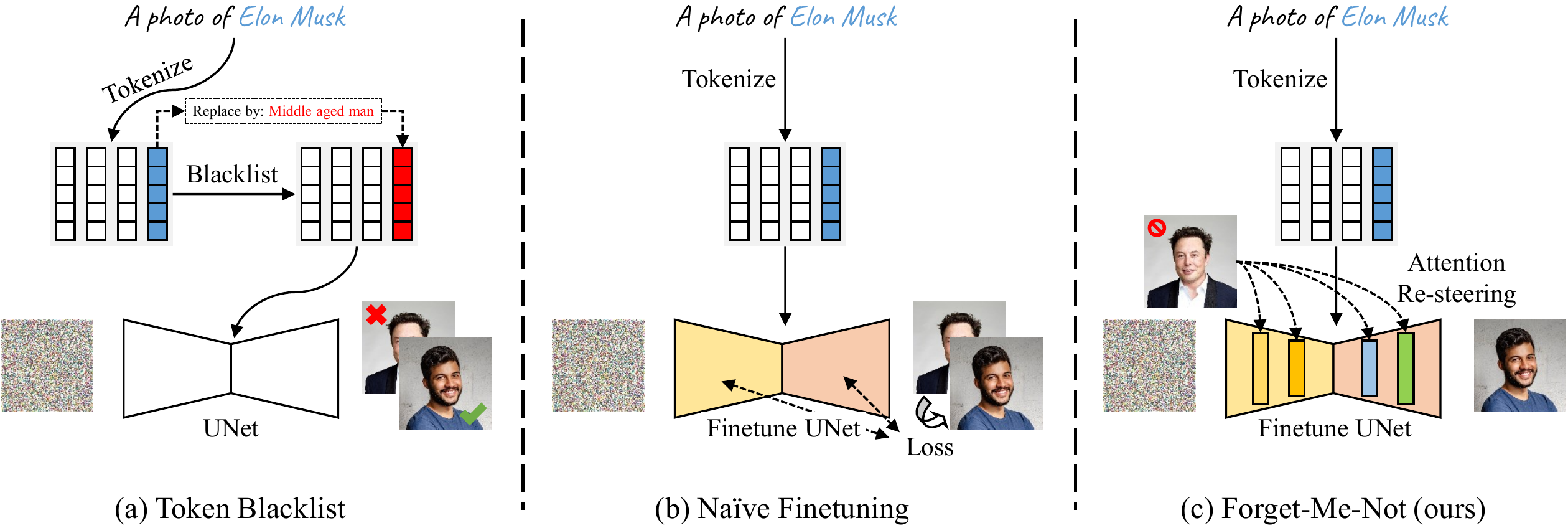}
    \caption{This figure shows two baseline forgetting methods and our proposed Forget-Me-Not. The target concept to forget is Elon Musk. One baseline is (a) Token Blacklist that simply replaces the target token with a different one. The other baseline is (b) Naive Fintuning in which instead of replacing tokens, it finetunes model weights so that the new weights generate outputs containing unrelated concepts. Our method (c) Forget-Me-Not utilizes Attention Re-steering in which we finetune only UNet to minimize each of the intermediate attention maps associated with the target concepts to forget. }
    \label{fig:forgetting_methods}
\end{figure*}

\section{Related Works}

\subsection{Text-to-Image Synthesis}

Image generation has been a challenging but very attractive research area, whose goal is to synthesize natural-looking images. In the past decade, we have witnessed the rapid advance of it from unconditional generative models to conditional generative models with powerful architectures of auto-regressive model~\cite{dalle1, parti}, GAN~\cite{ic-gan,stylegan, stylegan3, stylenat, singan} and diffusion process~\cite{ddpm, ddim, A-Fast-ODE-Solver-for-Diffusion, GENIE-diffusion, eDiff-I-diffusion, Score-Based-Generative-Modeling-diffusion}. Early works focus on unconditional, single-category data distribution modeling , such as hand-written digits, certain species of animals, and human faces~\cite{deng2012mnist, afhq, stylegan, celeba}. Though, unconditional models quickly achieves photo realistic results among single-category data, it's shown that mode collapsing issue usually happens when extending data distributions to multiple-category or real image diversity~\cite{ic-gan, Unrolled-GAN, Wasserstein-GAN}.

To tackle the model collapsing problem, the conditional generative model has been introduced. Since then, different types of data have been used as the conditioning for generative models, e.g. class labels, image instances, and even networks~\cite{ic-gan, Conditional-gan} etc. At the same time,  CLIP~\cite{clip, Ilharco_OpenCLIP_2021}, a large-scale pretrained image-text contrastive model, provides a text-image prior of extremely high diversity, which is discovered to be applicable as the conditioning for generative model~\cite{glide, vqgan-clip, more-control-for-free}. Nowadays, DALL-E~2 ~\cite{dalle2} and Stable Diffusion~\cite{Rombach_2022_CVPR} are capable of generating high quality images solely conditioning on free-form texts, inheriting the diversity of billions of real images scraping from the Internet. Subsequently, a line of work seeks to efficiently adapt the massive generative model to generate novel rendition of an unseen concept represented by a small reference set, leveraging the great diversity. Dreambooth~\cite{ruiz2022dreambooth} proposed to adapt the model by finetuning all of its weights, while it requires enormous storage to save newly adapted weights. Textual Inversion~\cite{gal2022textual} and LoRA~\cite{lora} ameliorate the issue by adapting the model by adding a small set of extra weights.

\subsection{Model Unlearning}

However, this great diversity comes at a price. It incurs potential risk of privacy leakage and copyright infringement.~\cite{Extracting-training-data-from, diffusion-art-or-digital-forgery} have successfully retrieved samples from Stable Diffusion that are highly faithful to real training examples. Therefore, being able to forget/unlearn certain concept in a model without hurting the generative ability for the rest is of both research and practical interests. Similar topics have been seen in fields other than conditional generative modeling. In model-agnostic meta-learning,~\cite{learn-to-forget-for-meta-learning} noted selectively forgetting the influence of prior knowledge in a network improves the performance in adapted tasks.~\cite{learning-to-unlearn-instance-wise-unlearning-for-pretrained-classifiters, Learn-to-Forget-Machine-Unlearning, Certifiable-Machine-Unlearning-for-Linear-Models, Zero-Shot-Machine-Unlearning} explores the unlearning of a set of requested data points in a pretraind model.

Our work differs from existing forgetting and unlearning works in a few aspects. First, we study forgetting in the context of text-to-image generative models. Second, we are deleting not only requested data points represented by a small reference set, but the concept behind those data points, which possesses significant impact in text-to-image generation due to the fact that it's almost impossible to enumerate all prompts and synonyms relating to a concept.


\section{Method}
\subsection{Preliminaries}

\textbf{Diffusion models}~\cite{ddpm, ddim, dmbeatgan} are denoising models that iteratively restore data $x_0$ from its Gaussian noise corruption $x_T$ with a total step number $T$. Such a restoration process is usually known as the reverse diffusion process $p_{\theta}(x_{t-1}|x_t)$ and the opposite of the reverse process is the forward diffusion process that blends the signal with noise $q(x_{t}|x_{t-1})$:

\vspace{-0.5cm}
\begin{equation*}\begin{aligned}
    q(x_t \vert x_{t-1}) &= \mathcal{N}(x_t; \sqrt{1 - \beta_t} x_{t-1}; \beta_t\mathbf{I}) 
    \\
    \\
    p_\theta(x_{t-1} \vert x_t) &= \mathcal{N}(x_{t-1}; \mu_\theta(x_t, t); \Sigma_\theta(x_t, t))
\label{eq:frdp}
\end{aligned}\end{equation*}

\noindent Both forward and reverse processes are presumably Markovian chains, so we can express the likelihood of both processes as:

\vspace{-0.5cm}
\begin{equation*}\begin{aligned}
    q(x_{1:T} \vert x_0) &= \prod^T_{t=1} q(x_t \vert x_{t-1})
    \\
    p_\theta(x_{0:T}) &= p(x_T) \prod^T_{t=1} p_\theta(x_{t-1} \vert x_t) 
\end{aligned}\end{equation*}

\noindent The loss function for the diffusion process is then to minimize the variational bound $\mathcal{L}_{vlb}$ of the negative log-likelihood $p_{\theta}(x_0)$ (\ie maximize the likelihood of $x_0$ as the final denoised result from a model with parameters $\theta$):

\begin{equation*}\begin{aligned}
    \mathcal{L}_\text{VLB} 
    &= \mathbb{E}\left[-\log p_{\theta}(x_0)\right] 
    \\    
    &\le \mathbb{E}_q \left[ -\log\frac{p_\theta(\mathbf{x}_{0:T})}{q(\mathbf{x}_{1:T}\vert\mathbf{x}_0)} \right]
\end{aligned}\end{equation*}
\vspace{0.3cm}

\textbf{Cross-Attentions}~\cite{attention-is-all-your-need} are widely adopted deep learning modules used in discriminative models~\cite{vit, detr, oneformer}, conditional generation models~\cite{dalle2, imagen, Rombach_2022_CVPR} as well as language models~\cite{bert, t5, flant5}. The purpose of cross-attention is to transfer information from conditional inputs to hidden features through dot product and softmax. For example, in stable diffusion~\cite{Rombach_2022_CVPR}, the hidden feature serves as the query $Q$ and context serves as key $K$ and value $V$. Assume $Q$ and $K$ has dimension $d$ for inner product, the output $h$ is then computed as the following:

\begin{equation*}\begin{aligned}
    h = \text{softmax}(\frac{QK^T}{\sqrt{d}})V \\
\end{aligned}\end{equation*}

\noindent It is important to note that such $QKV$ assignments are not fixed. Other assignments, such as conditional-driven queries and feature-driven keys and values, may also have their usage.

\begin{figure}[t]
    \centering
    \includegraphics[width=0.9\columnwidth]{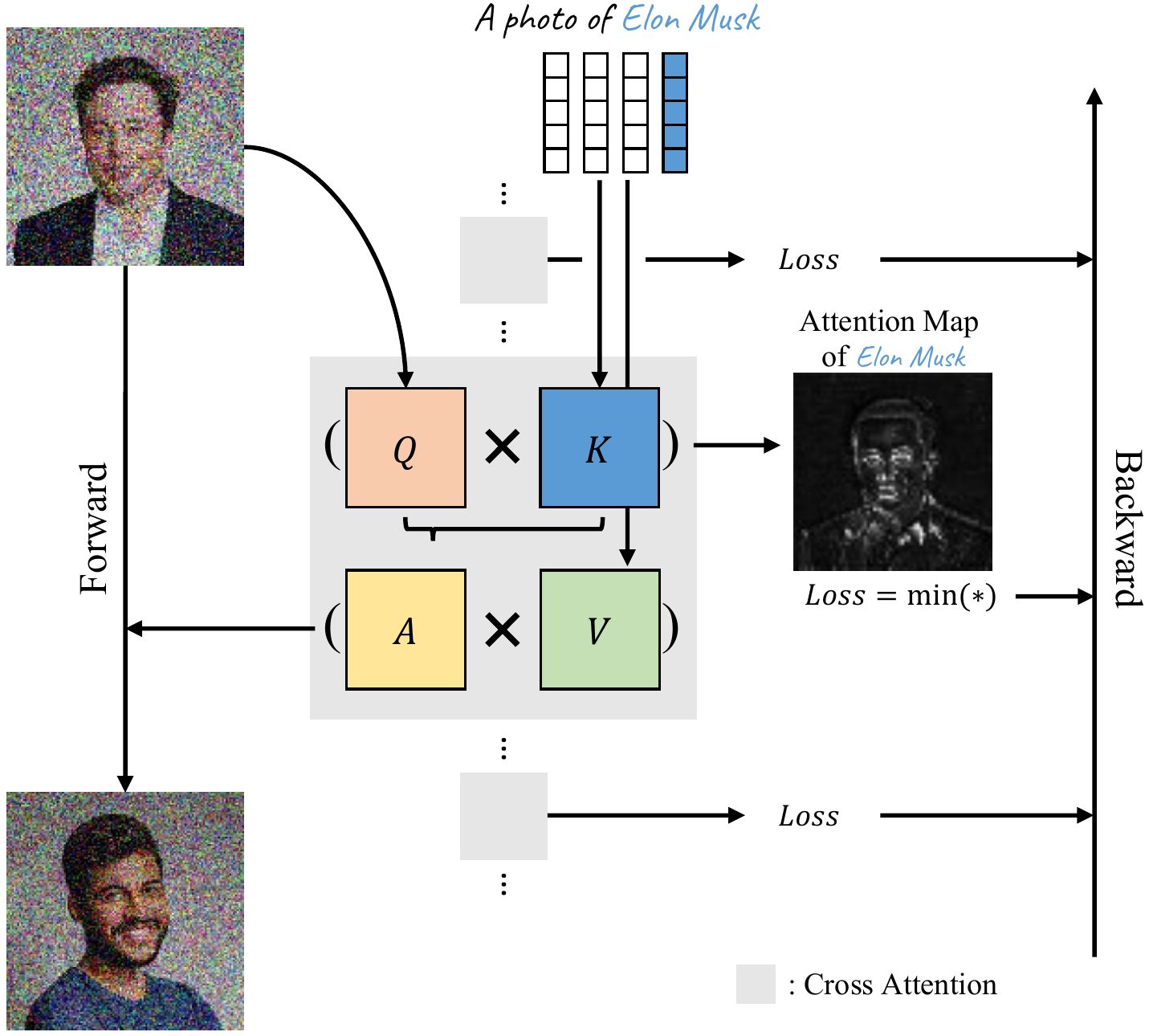}
    \caption{This figure shows the Attention Re-steering we proposed in our Forget-Me-Not method, in which we set the objective function to minimize the attention maps of target concepts (\ie Elon Musk in this case) and correspondingly finetune the network.}
    \label{fig:attn_resteering}
    \vspace{-3mm}
\end{figure}

\subsection{Concept Forgetting}\label{sec:concept-forgetting}

A concept is an abstract term representing an intuited object of thought, which also serves as the foundation for people's perceptions. Specifically for computer vision, we may recognize concepts as tangible things, including identities, objects that physically existed, style of images, object relations, and even poses and behavior. 
Concept forgetting, literarily speaking, is the action of reverting a model from understanding certain concepts. 
On contrast to machine unlearning, which aims to delete the fields around designated data points, we define concept forgetting in diffusion models as the disentanglement of concept prompts and visual contents. This definition allows models to retain their generative abilities to the greatest extent possible.

Besides, we set the following four goals for concept forgetting research: 

\begin{itemize}
    \item \textbf{Performance: } the proposed approach should at best remove target concepts from the model.
    \item \textbf{Integrity: } the proposed approach should at best keep other concepts of the model. 
    \item \textbf{Generality: } the proposed approach can be applied to a wide range of concepts that covers all aspects of human perceptions.
    \item \textbf{Flexibility: } the proposed approach can be applied to various models of different tasks and domains.
\end{itemize}

\subsection{Forget-Me-Not}\label{sec:forget-me-not}

To fulfill the end goals we mentioned in Section~\ref{sec:concept-forgetting}, we introduce \textbf{Forget-Me-Not}, a heuristically and important approach toward ultimate concept forgetting. One highlight of Forget-Me-Not is that it addressed all goals to some extent. Forget-Me-Not is well-capable in removing a wide variety of concepts without manipulating too much on other outputs. Its underlying methodology, \textit{attention resteering}, fits almost all major text-to-image models and may extend to other conditional multimodal generative models. Besides, Forget-Me-Not is a practical solution for many models, all thanks to its low-cost and plug-and-play design. 

\textbf{Baselines:} Before we dive deep into Forget-Me-Not, one may notice that simple approaches such as Token Blacklisting and Naive Finetuning can as well be workable solutions for text-to-image models (see Figure~\ref{fig:forgetting_methods}). Thus we use these approaches as the baselines of this work. Token Blacklisting wipes out the token embeddings to forget concepts, like crossing out the dictionary index. It is an instant solution, but one may still recover the forgotten concept via token inversion; therefore, this approach doesn't forget anything in actuality. Additionally, blacklisting a specific token may inadvertently affect other concepts that share the same prompt, making it challenging to target a specific concept without affecting others. Though this side effect can be mitigated through prompt engineering, it is still difficult to eliminate due to the variability of natural language.
Another baseline, Naive Fintuning, deliberately corrupts the target concept by finetuning the model so that the target concepts are remapped on random images. The downside of this approach is obvious: it breaks model integrity by simultaneously corrupting other unrelated ideas during its finetuning process. As a summary, these pros and cons are listed in Table~\ref{table:baseline_pandc}.

\begin{table*}[h!]
\centering
\resizebox{\textwidth}{!}{
    \begin{tabular}{
            L{3cm}
            C{3cm}
            C{3cm}
            C{3cm}
            C{3cm}}
        \toprule
            Methods
            & Performance & Integrity
            & Generality & Flexibility
            \\
        \midrule
        \midrule
        Token Blacklisting
            & \color{cons}{No forgetting}
            & \color{neutral}{Inevitably affects other concepts sharing overlapping prompts}
            & \color{neutral}{Within the vocabulary of the tokenizer}
            & \color{neutral}{Tokenizer required}
            \\
        \midrule
        Naive Finetuning
            & \color{pros}{Successfully removes concept}
            & \color{cons}{Removes unrelated concept by fault}
            & \color{pros}{Applies to any concepts with sufficient data.}
            & \color{pros}{Applies to any models}
            \\
        \midrule
        Forget-Me-Not
            & \color{pros}{Successfully removes concept}
            & \color{pros}{Maintains most of the model's integraity.}
            & \color{pros}{Applies to any concepts with few data samples}
            & \color{neutral}{Only applies to models with cross attention}
            \\
        \bottomrule
    \end{tabular}
}
\vspace{0.3cm}
\caption{
    This table compares pros ({\color{pros}{green}}) and cons ({\color{cons}{red}}) on the four major aspects of concept forgetting between baselines and the proposed Forget-Me-Not. If an approach can handle an aspect to some extend, the corresponding explanation is marked in {\color{neutral}{yellow}}.
}
\label{table:baseline_pandc}
\end{table*}

\textbf{Attention Resteering:} One may notice that both baseline approaches fall into the ``performance or integrity" dilemma, while our Forget-Me-Not can elegantly resolve this challenge. The core method behind the scenes is the attention resteering loss design, which enables a precise backpropagation of those forgetting concepts. Figure~\ref{fig:attn_resteering} shows the diagram of this idea, in which we first locate the context embeddings associated with the forgetting concept; compute the attention maps between input features and these embeddings; then minimize the attention maps and backpropagate the network. Such attention resteering can be plugged into any cross-attention layers of the network. It also decouples model's finetuning from its original loss functions(\ie variational lower bound for diffusion models), making the process a much simpler solution. One of our main focuses in this work is concept forgetting of text-to-image models; therefore we carry out attention resteering on all cross-attention layers of UNet~\cite{unet} in Stable-Diffusion (SD)~\cite{Rombach_2022_CVPR}, which yields the best performance on most of the concepts. The entire algorithm of Forget-Me-Not can be found in Algorithm~\ref{alg:forget_me_not}.

\begin{algorithm}
\caption{Forget-Me-Not on diffuser}
\begin{algorithmic}[1]
\Require Context embeddings $\mathcal{C}$ containing the forgetting concept, embedding locations $\mathcal{N}$ of the forgetting concept, reference images $\mathcal{R}$ of the forgetting concept, diffuser $G_\theta$, diffusion step $T$.

\Repeat 
\State $t \sim \text{Uniform}([1 \dots T]); \epsilon \sim \mathcal{N}(\mathbf{0}, \mathbf{I})$
\State $r_i \sim \mathcal{R}; c_j, n_j \sim \mathcal{C}, \mathcal{N}$ 
\State $x_0 \gets r_i$ 

\State $x_t \gets \sqrt{\bar{\alpha}_t}x_0 + \sqrt{1 - \bar{\alpha}_t}\epsilon$ 
\State \Comment{$\bar{\alpha}_t$: noise variance schedule}
\State $x'_{t-1}, A_t \gets G_\theta(x_t, c_j, t)$
\State \Comment{$A_t$: all attention maps}
\State $\mathcal{L} \gets \sum_{a_t \in A_t} \|a_t^{[nj]} \|^2$
\State \Comment{$\mathcal{L}$: attention resteering loss}
\State $\theta \gets \theta - \nabla_\theta \mathcal{L}$
\Until Concept forgotten
\end{algorithmic}
\label{alg:forget_me_not}
\end{algorithm}

\textbf{Optional Concept Inversion:} Although we may directly obtain context embeddings using prompts for most text-to-image models, this is not a generalized case for all concepts, particularly when a) the forgetting concept is out of vocabulary; b) the model doesn't have a vocabulary; c) the description of the forgetting concept is unclear. To overcome the challenge, we optionally include the textual inversion~\cite{gal2022textual} as a fixed overhead before Forget-Me-Not to strengthen its generality on all concepts. In practice, we also notice that such inversion helps text-to-image models more precisely identify the forgetting concept and thus improves their performance. More results can be found in our Experiment Section.

\section{Experiments}
\subsection{ConceptBench}
To meet our need to evaluate Forget-Me-Not and potential future concept forgetting approaches, we introduce a benchmark, namely ConceptBench.

It is important to note that several existing benchmarks, such as ~\cite{lin2014microsoft, imagen}, help assess overall generation quality. However, none are specifically designed to measure a model's ability to memorize and forget. ConceptBench utilizes instances from LAION~\cite{laion}, forming three categories,identity, object, and style, ranging from discrete to abstract and easy to hard.

Identity refers to the unique and discrete features of each instance. Specifically, we examine identity in subcategories such as person, franchise, animal, and brand. The person and franchise subcategories are of particular interest due to potential privacy or copyright issues associated with generated images of celebrities or intellectual properties. For the animal breed subcategory, detailed instances may include specific breeds, such as ``Corgi" or ``Husky", which belong to the more general ``dog" category but have distinct visual features. In the case of brands, they represent abstract concepts of intangible objects that can manifest as logos throughout our daily lives.

Object is a broader concept encompassing multiple variations. For example, ``dog" refers to various breeds of dogs that share common features. By combining identity instances mentioned earlier, this category provides a hierarchical structure to examine the influence of concept forgetting on the model. We include food items like ``apple", ``banana", and ``broccoli", man-made objects such as ``airplane", ``keyboard", ``motorcycle", ``umbrella", and ``boat", and general animals like ``dog" and ``horse."

Style is an abstract concept that determines the overall appearance of generated images. ConceptBench incorporates styles such as ``Van Gogh", ``Picasso", ``doodle", ``pixel art", ``neon", and ``sketch."




\subsection{Baseline}

\begin{figure}[t]
    \centering
    \includegraphics[width=\columnwidth]{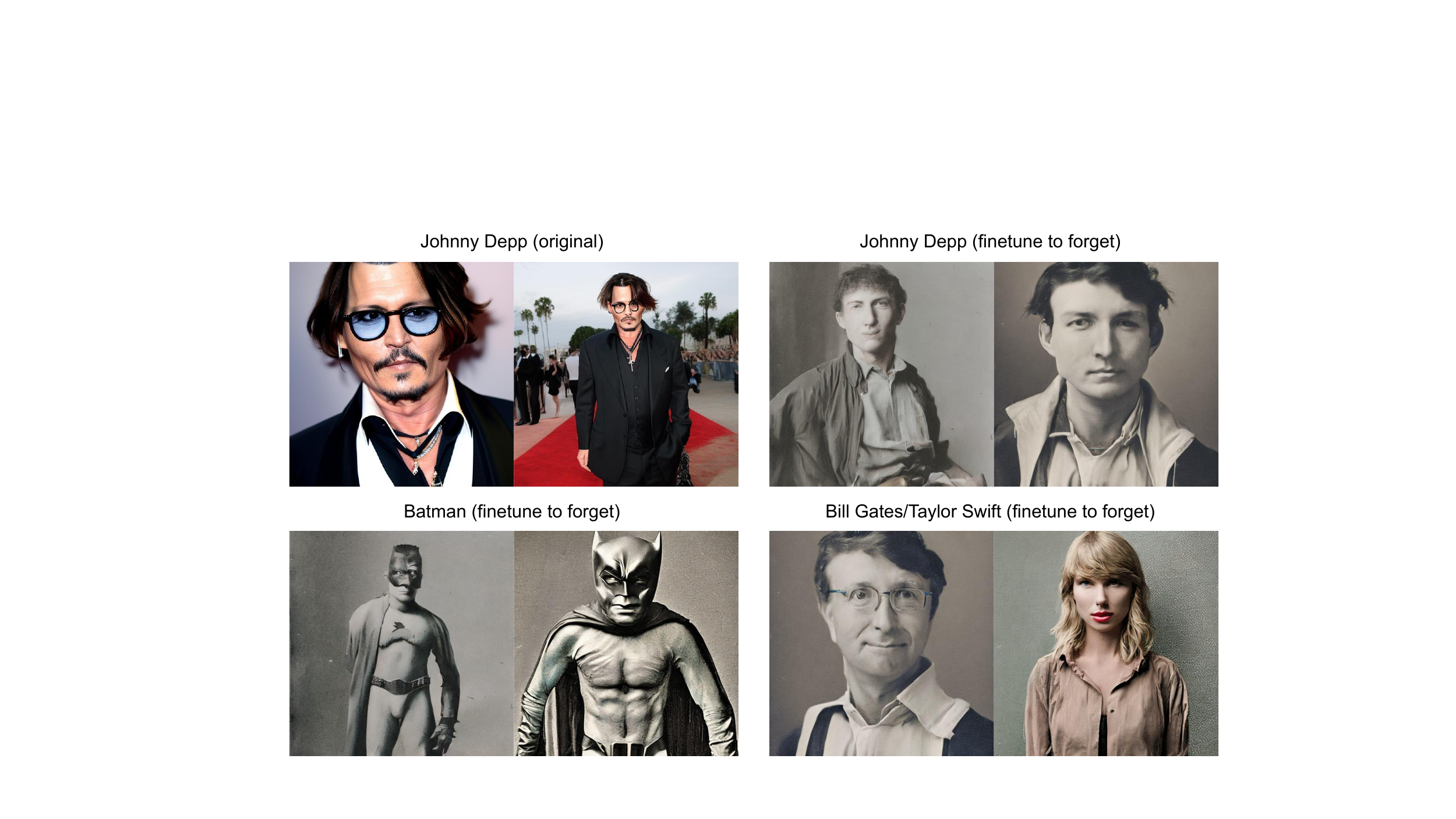}
    \caption{Finetuning to forget concept ``Johnny Depp" with unrelated images of ``a photo of man". This method distorts other concepts with visual details of selected unrelated images.}
    \vspace{-0.3cm}
    \label{fig:baseline}
\end{figure}

In view of the multi-component nature of Stable Diffusion models, there are several naive methods that can be used to superficially remove a concept from them, such as blacklisting keywords in prompts, removing specific tokens from the tokenizer dictionary, or tuning the model with unrelated images to divert the target concept, as illustrated in Figure \ref{fig:forgetting_methods}(a)(b). However, these methods can result in a significant deterioration and shifting of the model's generation capability. Removing tokens from the dictionary can alter the tokenization of prompts where those tokens were previously used and affect the generation of other prompts with overlapping tokens. For instance, removing tokens of ``Hillary Clinton" could lead to dysfunctionality in generating ``Bill Clinton". Naive finetuning to forget with unrelated images explicitly overwrites the visual representation of a concept with extra data and runs the risk of compromising existing concept space, as shown in Figure~\ref{fig:baseline}. Moreover, it is impossible to exhaust test all relation-based concepts for blacklisting or finetuning.


\begin{figure*}[h!]
    \centering
    \includegraphics[width=1\textwidth]{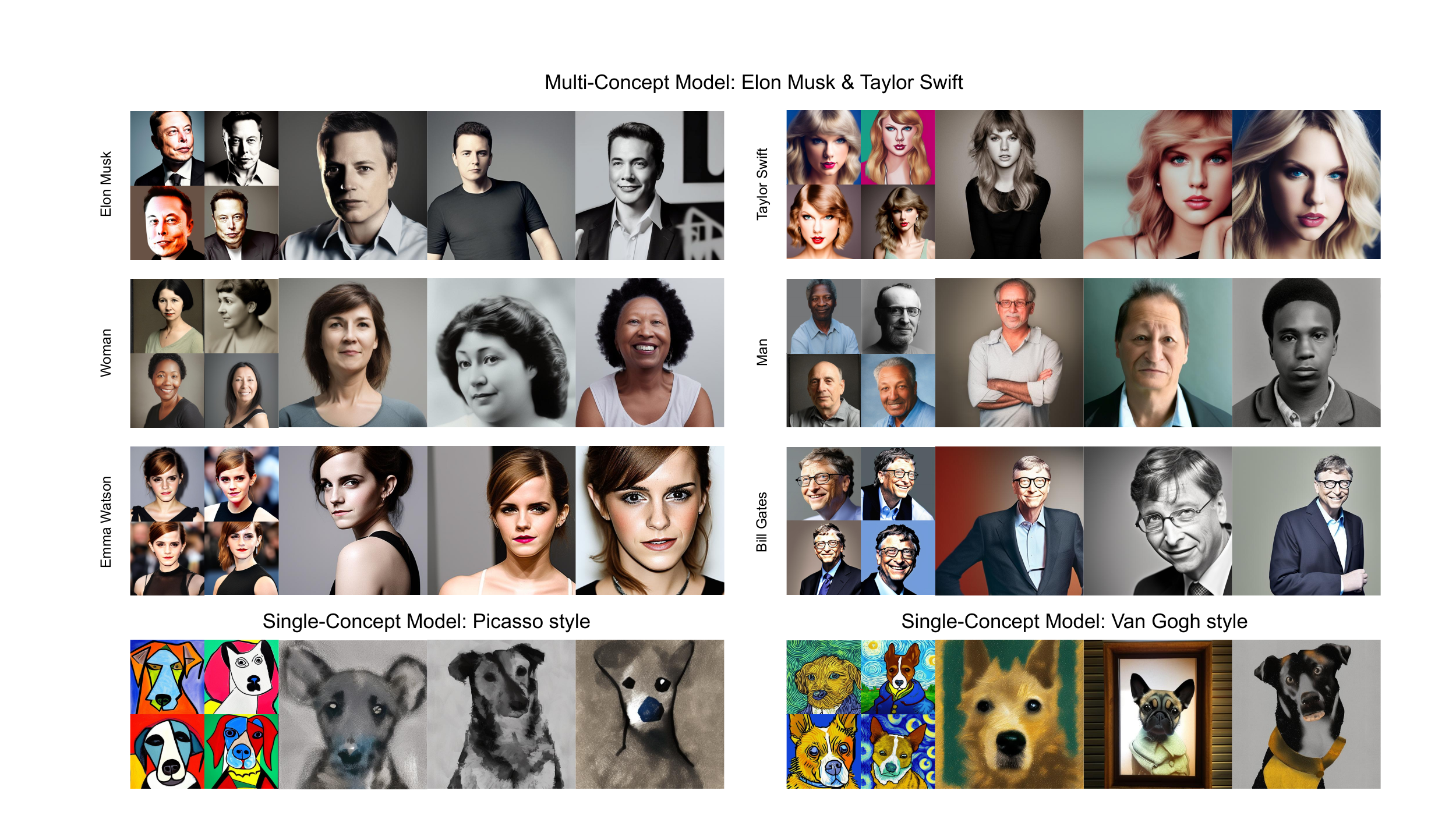}
    \caption{Results of concept forgetting using our method. The first 2x2 grid shows the original samples in Stable Diffusion. The subsequent 3 images are sampled after concept forgetting, using the same prompt.  The top 3 rows are from a multi-concept model targeting both Elon Musk and Taylor Swift, demonstrating the multi-concept forgetting capability. Control concepts such as Bill Gates and Emma Watson manifest that our approach has minimal impact on concepts other than target ones. The last row shows two single-concept model of styles. Output images were generated with prompts: ``a photo of X" (top 3 rows), ``a dog in X style" (bottom row).}
    \label{fig:forgetting-in-SD}
    \vspace{-3mm}
\end{figure*}


\begin{figure*}[h!]
    \centering
    \includegraphics[width=\textwidth] {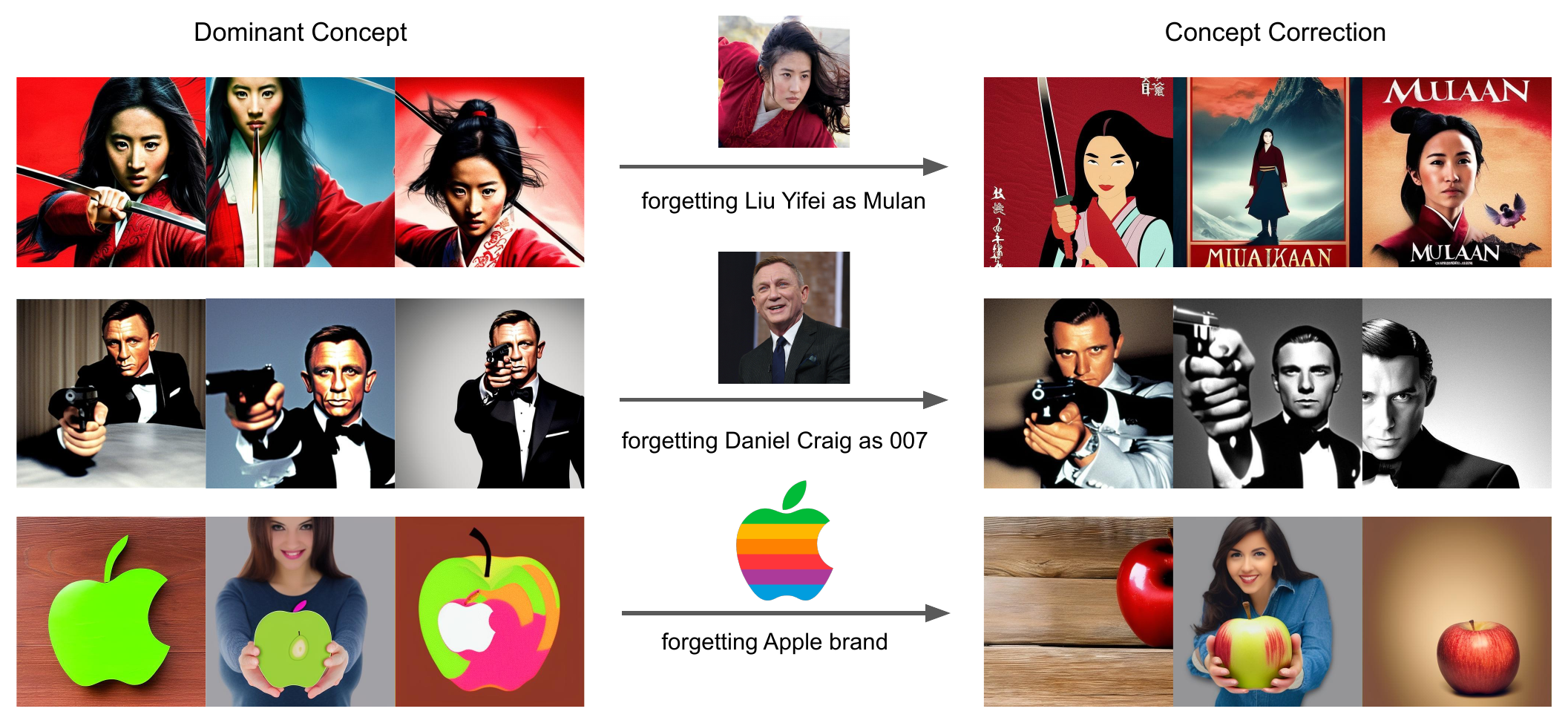}
    \caption{Concept Correction: Once the dominant concept has been diminished by our method, the lesser concepts of an semantic-rich prompt become more prominent in generated results. Output images were generated with prompts (top to bottom): ``a movie poster of Mulan", ``James Bond", ``apple shape".}
    \label{fig:semantic-diminishing}
\end{figure*}

\vspace{0.3cm}

\subsection{Qualitative Comparison}

We present the results of concept forgetting from our benchmark, illustrated in Figure~\ref{fig:forgetting-in-SD}, the Multi-concepts model of Elon Musk and Taylor Swift demonstrates our method's ability to perform multi-concept forgetting. As shown in the first row, both target concepts have been forgotten. We evaluated the impacts of forgetting specific concepts on other related concepts, examining four related concepts to Elon Musk and Taylor Swift - man, woman, Bill Gates, and Emma Watson. As shown, Forget-Me-Not achieved good content preservation and visual quality. However, we observed minor pose and style changes in man and Bill Gates. Based on these findings, we posit that our approach may have a greater impact on closely related concepts than on others. Additionally, the last row shows that a new painting style is emerging after forgetting Piccaso and Van Gogh styles.


\begin{figure}[h]
    \centering
    \includegraphics[width=\columnwidth]{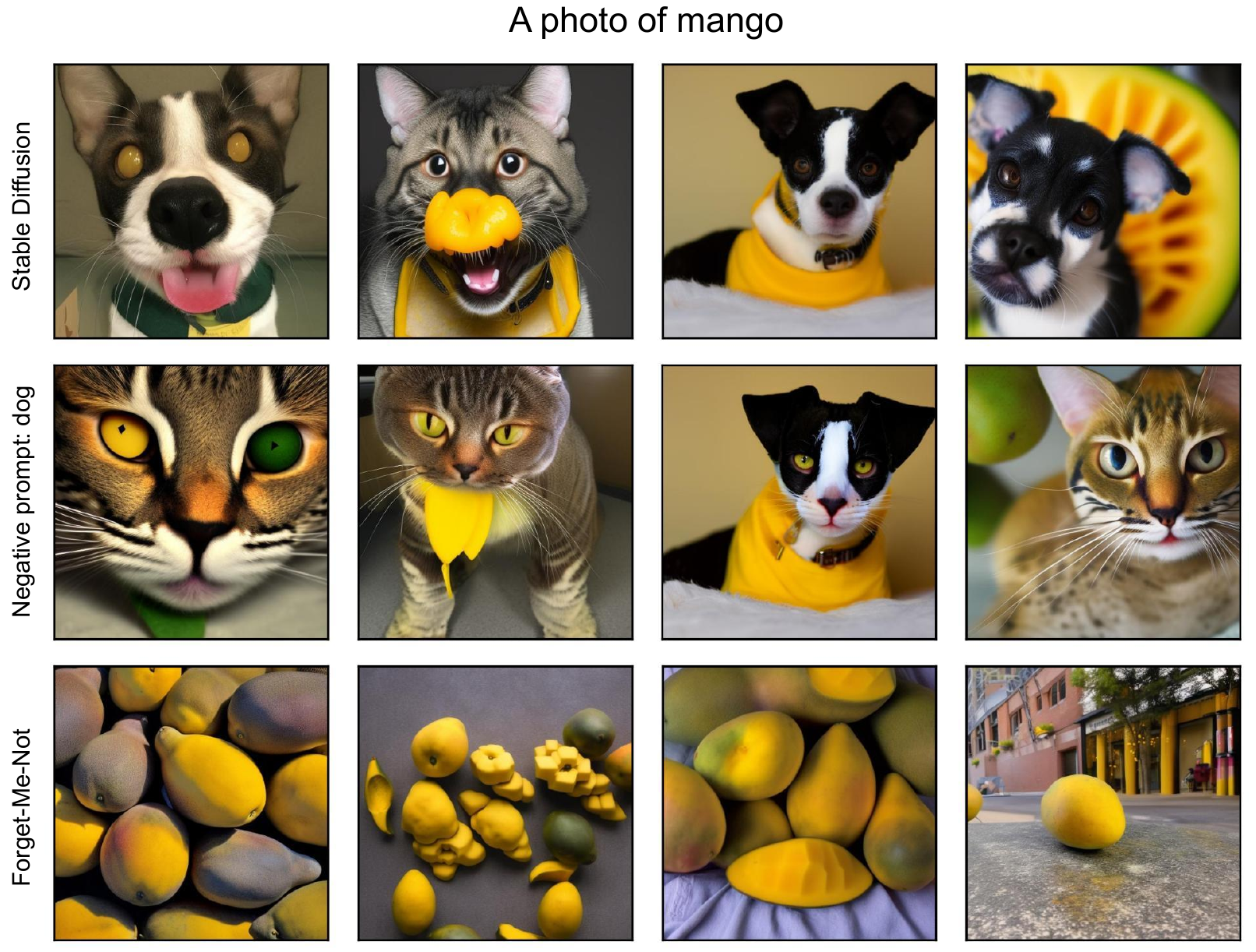}
    \caption{In concept correction, our method has the advantage of comprehensive forgetting over negative prompt.}
    \label{fig:comprehensive-forgetting}
    \vspace{-3mm}
\end{figure}

\subsection{Quantitative Analysis}\label{sec:quantitative-Analysis}


\textbf{Memorization Measurement} Textual inversion~\cite{gal2022textual} can be employed to identify the token embeddings that best correspond to images. We utilize this technique to measure the concept embedding changes of anchor images toward a reference before and after forgetting. These changes can be seen as generative model's memorization level of a concept, which we term Memorization Score.

In the case of ``Elon Musk", prompt ``Elon Musk" is used as reference. Its concept embedding ($\textbf{emb}_r$) is obtained by processing it through text encoder. Subsequently, we invert the same anchor images of Elon Musk using \textit{original model} and  \textit{forgetting model} respectively. Concept embeddings of anchor images are obtained by processing inverted tokens through text encoder, resulting in two sets: original textual inversion ($\textbf{emb}_o$) and forgetting textual inversion ($\textbf{emb}_f$).  Only the pooler tokens of concept embeddings are used for measurement. The change in concept embedding is quantified as the difference between $\cos (\textbf{emb}_r, \textbf{emb}_o)$ and $ \cos (\textbf{emb}_r, \textbf{emb}_f)$. A decrease indicates successful forgetting. 
Since the textual inversion process bring the randomness of embeddings, we compute the average Memorization Score over five running times. 
We present memorization scores 
from each sub-category in Table~\ref{table:memorization_score}. Additional results can be found in the Supplementary material.
\begin{table}[htbp]
\begin{center}
\begin{tabular}{|c|c|c|}
   \hline
   Concept  & \begin{tabular}{@{}c@{}}Initial \\ Mem Score\end{tabular} & \begin{tabular}{@{}c@{}}Forgetting \\ Mem Score\end{tabular} \\
   \hline
    Elon Musk  & 0.943 & 0.848 \\
    Mickey Mouse  & 0.948 & 0.836  \\
    Zebra & 0.972 & 0.899 \\
    Google & 0.940 & 0.811\\
    Apple & 0.696 & 0.493 \\
    Horse & 0.877 & 0.808 \\
    Van Gogh  & 0.916 & 0.684\\
   \hline
\end{tabular}
\caption{Memorization Scores of instances from each sub-categories.}
\label{table:memorization_score}
\end{center}
\vspace{-8mm}
\end{table}







\subsection{Concept Correction}

It has been observed that in text-to-image models, the semantics of a prompt are often dominated by the one with the most number of image-text examples in the training set, resulting in the suppression of inferior semantics during inference. Figure~\ref{fig:semantic-diminishing} exemplifies this scenario, where generation is dominated by a concept that is strongly correlated with a prompt due to unbalanced training examples. In the case of the James Bond series, the generation results are overwhelmingly dominated by Daniel Craig, as shown in the middle row. However, our method manages to diminish the most prominent semantic in the prompt, i.e., Daniel Craig, and make other James Bond actors visible. Similarly, in the case of Mulan series and the homonym of ``apple", where the apple fruit and Apple brand are competing with each other, our method successfully corrects target concept in generated images .

Negative prompt is a technique used in text-to-image synthesis to eliminate unwanted concepts associated with a prompt. However, their use can result in negative impacts on other aspects of the image, such as changes to its structure and style. Furthermore, negative prompts fail to correct undesirable concepts under certain circumstances. For example, in Figure~\ref{fig:comprehensive-forgetting}, ``a photo of a mango" consistently generates dog images. This is because the name ``mango" is commonly used as a pet name for dogs, and people upload photos of their dogs to the internet, which are collected as training data. In this case, using a negative prompt for dogs would be ineffective, as mango is also a popular cat name, creating the problem of endlessly expanding negative prompts. However, our method successfully brings back the mango fruit by forgetting the connection between ``a photo of mango" and dog/cat images.

\subsection{NSFW Removal}

In this section, we examine the effectiveness of our method in a real case of removing harmful contents. NSFW is an internet shorthand for ``not safe for work", used for indicating contents that are not wished to be seen in the public. Such content may include material even offensive for adult audiences. 
However, they inevitably present in large datasets such as LAION~\cite{laion}, even though NSFW detector has been used~\cite{laion-nsfw-checker}. Stable Diffusion, trained on LAION, is known for generating NSFW content when prompted with certain triggers.

To evaluate our method, we use a well-known NSFW-triggering prompt, ``a photo of naked" in Stable Diffusion v2.1 model. Using EulerAncestralDiscreteScheduler, inference-step 50, and scale-guidance 8, the model consistently generates images containing nude individuals. We use eight generated NSFW images as input for training Forget-Me-Not.

\begin{figure}[t]
    \centering
    \includegraphics[width=\columnwidth]{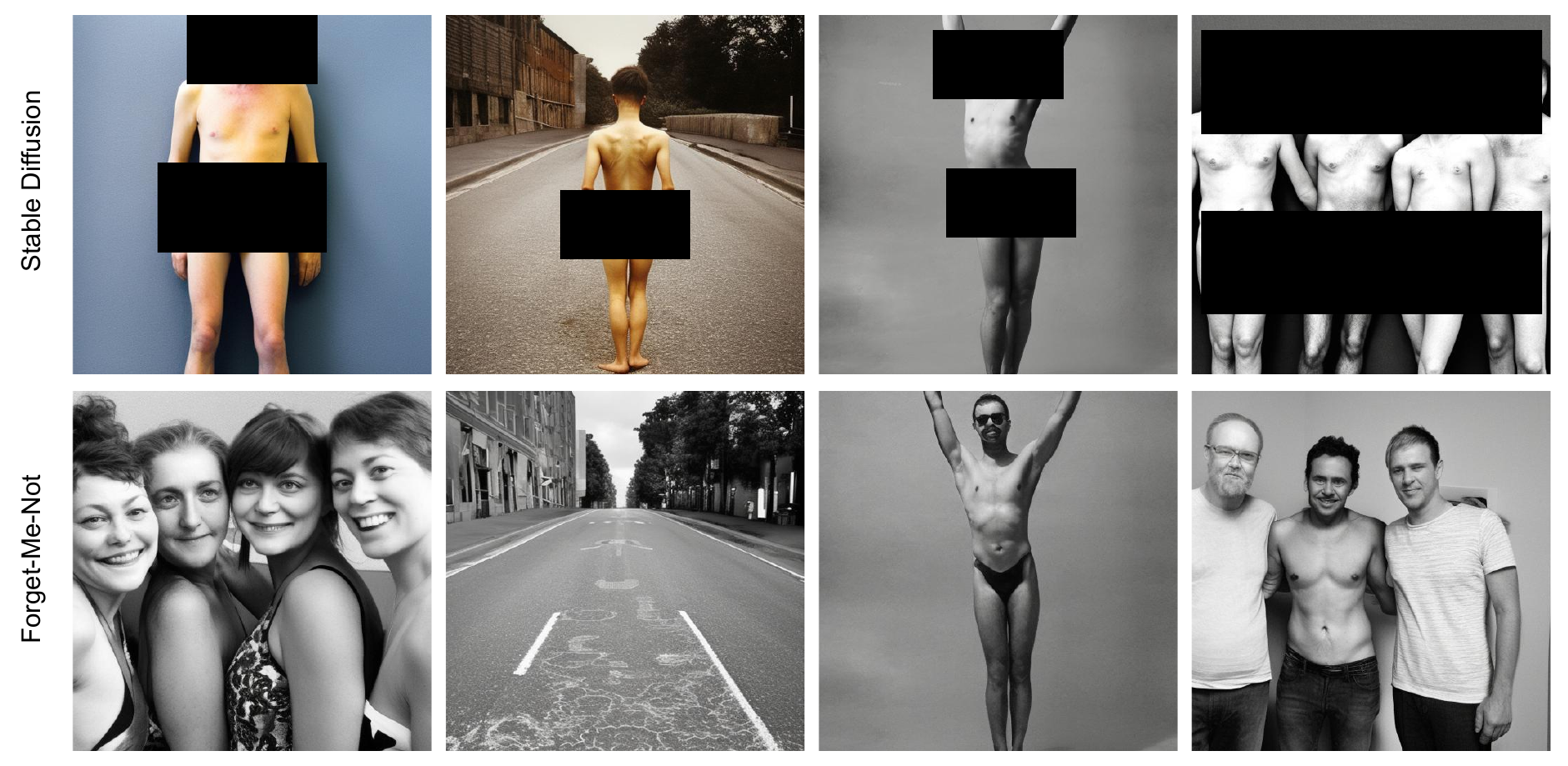}
    \caption{Results of removing NSFW contents triggered by ``naked". Faces and sensitive parts are blacked out. }
    \label{fig:nsfw}
    \vspace{-0.3cm}
\end{figure}

The results, shown in Figure~\ref{fig:nsfw}, indicate that the concept of ``naked" has been successfully forgotten. Notably, the second row shows that all sensitive visual cues have been changed in different ways. The first example changes abruptly from a naked man to a group of smiling women. In the second example, NSFW individual has been removed from the scene. The last two examples render clothed people, making them safe. Overall, our method achieved efficient forgetting of NSFW content without the need for additional data or the assistance of third-party NSFW detectors.

\subsection{Ablation Studies}

\begin{figure}[t]
    \centering
    \includegraphics[width=\columnwidth]{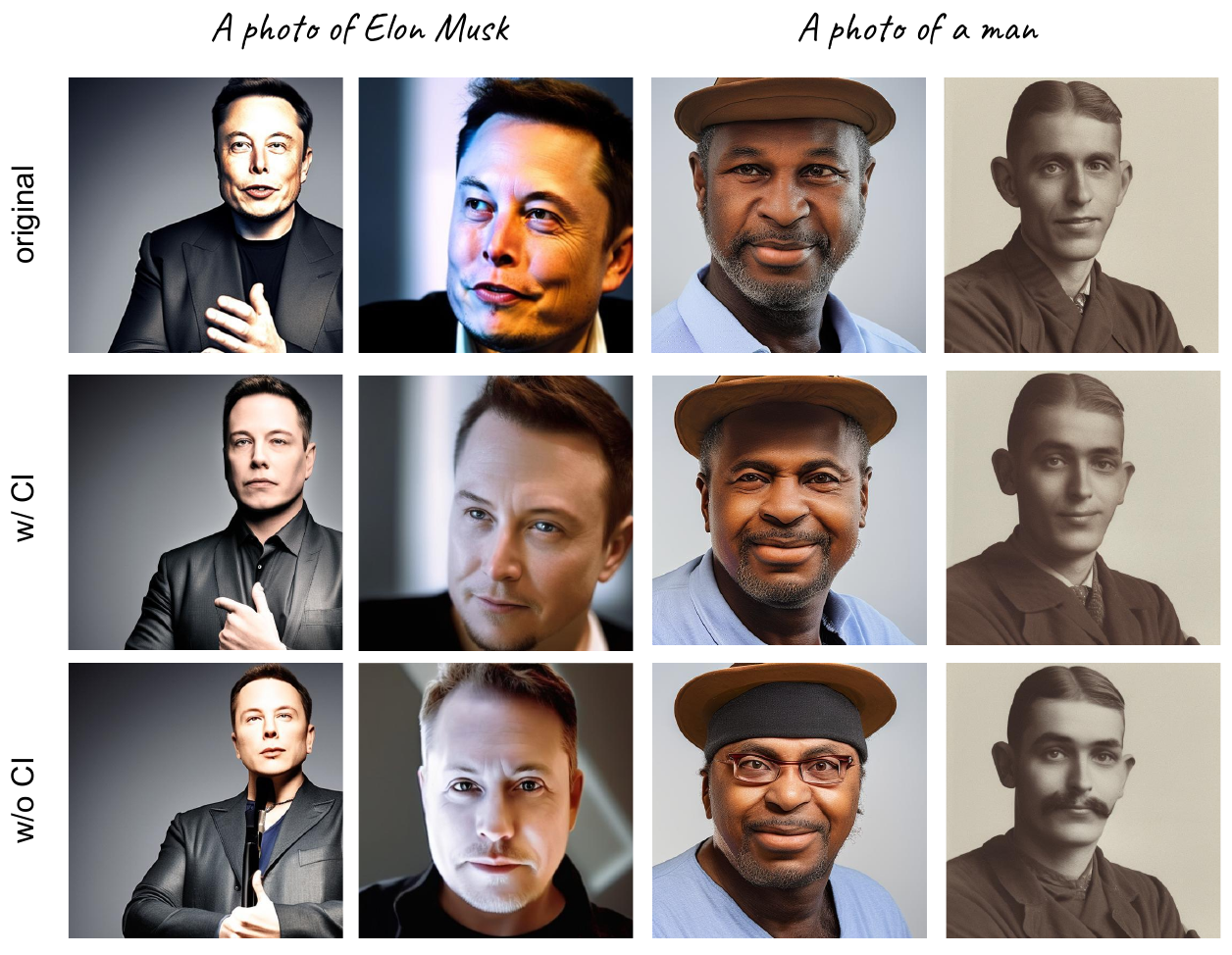}
    \caption{Improving fidelity to original model with concept inversion. Concept prompt tend to have diverse semantics, resulting in distortion in concept forgetting. CI extracts precise semantics into dedicated tokens, allowing for more pose and feature consistency.}
    \label{fig:ablation-ci}
    \vspace{-0.3cm}
\end{figure}

\textbf{Concept Inversion Ablation}\label{sec:ci-ablation}
We conducted experiments on concepts from ConceptBench, with and without using concept inversion (CI). Concept inversion is used to handle concepts that are difficult to describe using prompts. Generally, it can help extract the target concept from the prompt, resulting in more precise embeddings. However, precise embeddings may not be always ideal, see Section \ref{sec:concept-ablation} Concept Ablation. Our results show that CI can achieve higher fidelity for concepts that can be well-described in a prompt, as illustrated in Figure \ref{fig:ablation-ci}, where the model trained with CI preserved more of the original poses and details.

\begin{figure}[t]
    \centering
    \includegraphics[width=\columnwidth]{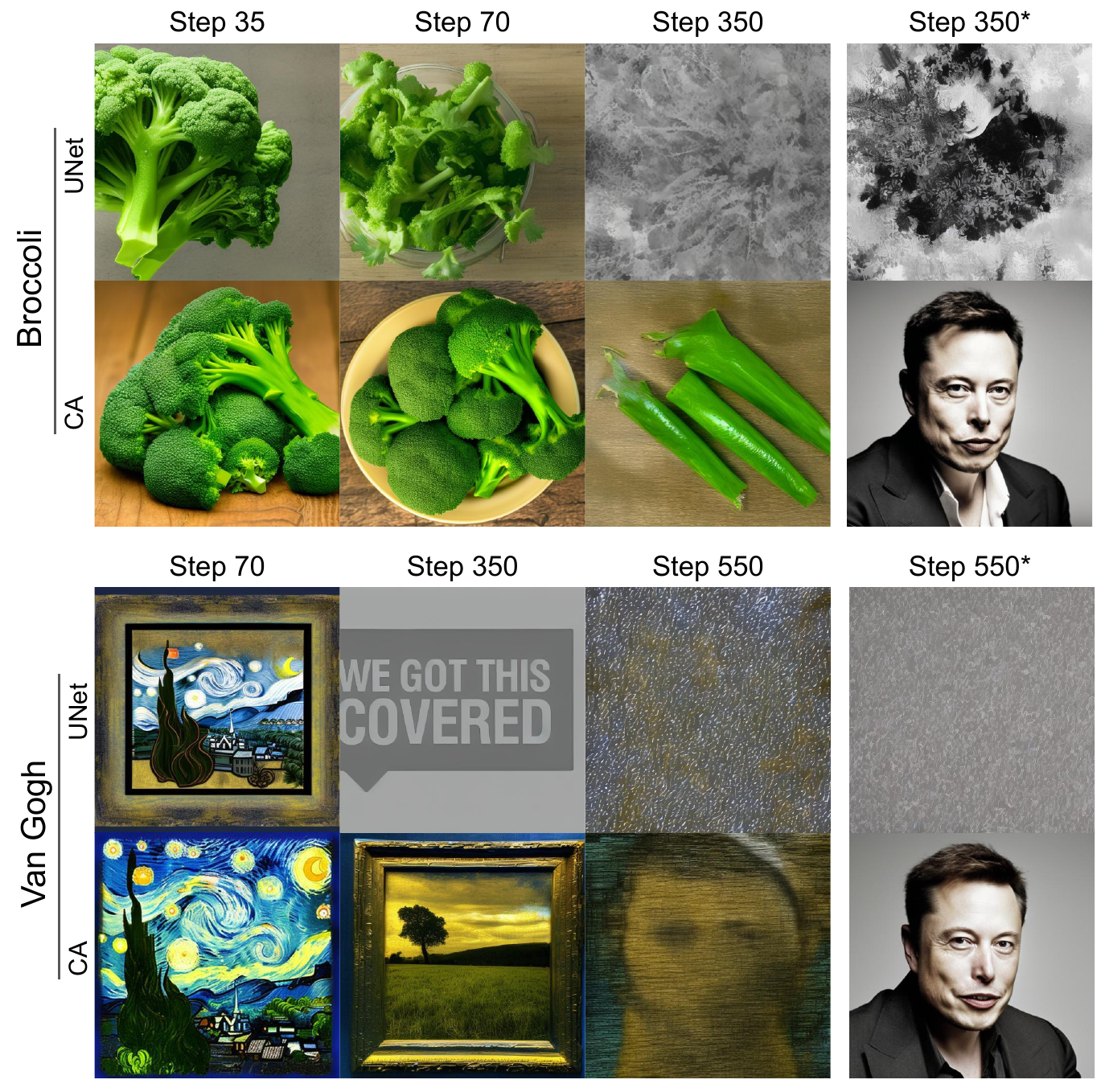}
    \caption{Trainable weights ablation using UNet and Cross Attention (CA). Compared to CA, UNet is more sensitive to optimization steps. The last column with Step X\textsuperscript{*} shows the control concept Elon Musk at Step X.}
    \label{fig:ablation-ca}
    \vspace{-0.3cm}
\end{figure}

\textbf{Trainable Weights Ablation}
We conducted experiments to compare finetuning the entire UNet model versus only finetuning the cross-attention (CA) layers. Cross-attention is a critical component in text-to-image generation, as it injects textual information into the image formation process. Given the same hyper-parameter settings except for steps, our results show that both methods can successfully achieve concept forgetting. However, finetuning the entire UNet model tended to break the model's generation capability in fewer steps. In some cases, the model collapsed before the forgetting process was complete, as show in the ``Broccoli" case of Figure \ref{fig:ablation-ca}.

\begin{figure}[t]
    \centering
    \includegraphics[width=\columnwidth]{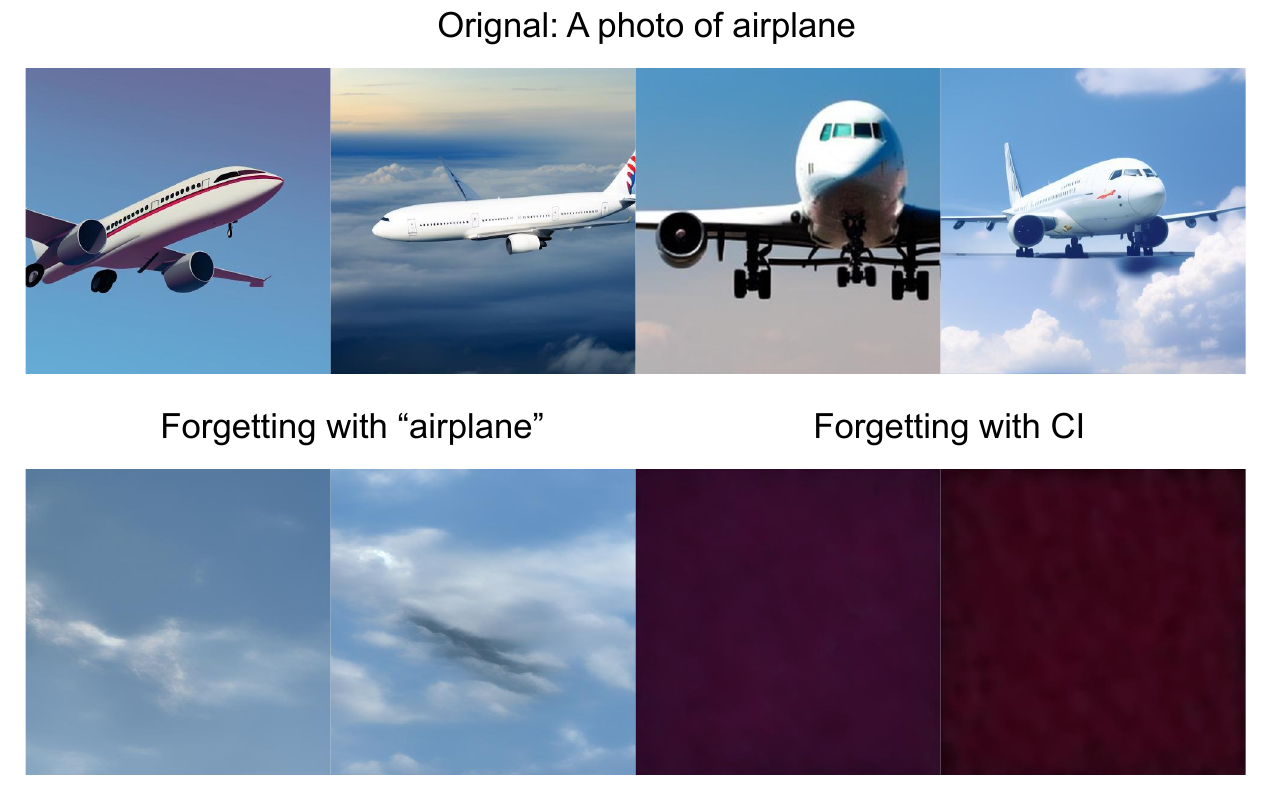}
    \caption{Comparison of different token embeddings in concept forgetting. Given concept airplane, we compare forgetting with either tokens of ``airplane" or inverted tokens using concept inversion, where forgetting with CI fails.   }
    \label{fig:ablation-concept}
    \vspace{-0.3cm}
\end{figure}

\textbf{Token Embedding of Concept Ablation} \label{sec:concept-ablation}
Our method relies on token embeddings of a concept, which are critical for computing the attention re-steering loss. As shown in Section \ref{sec:ci-ablation} on Concept Inversion Ablation, changing the token embeddings of a concept produces varying results. In Figure \ref{fig:ablation-concept}, we demonstrate a situation where concept prompt prevails concept inversion. By using the same settings except for token embeddings, prompt of ``airplane" succeeds while inverted tokens fails. We hypothesize that minimizing cross attention over these specific inverted tokens of ``airplane" tends to break generative capability of the model quickly.

\section{Conclusion}

In this study, we investigate concept forgetting in text-to-image generative models and introduce Forget-Me-Not. This lightweight approach enables ad-hoc concept forgetting using only a few either real or generated concept images; it can also be easily distributed using model patches. Forget-Me-Not is further naturally extended to enable concept correction and disentanglement. Our experiments demonstrate that Forget-Me-Not is successful in diminishing and correcting target concepts in Stable Diffusion. Additionally, we introduce ConceptBench and Memorization Score as evaluation metrics. Overall, our work provides a foundation for further research on concept forgetting and manipulation in text-to-image generation, and can be further extended to other conditional multimodal generative models to improve the accuracy, inclusion and diversity of such models.

\section{Social Impact \& Limitations}
\textbf{Social Impact} Our research has a positive social impact by offering an effective and cost-efficient method to remove and correct harmful and biased concepts in text-to-image generative models. These models are rapidly becoming the backbone of popular AI art and graphic design tools, used by a growing number of people. Our method can generate lightweight model patches that can be conveniently distributed to text-to-image model users like how conventional software patch works. Thus, our research takes a small step towards promoting fairness and privacy protection in AI tools, ultimately benefiting society as a whole.

\textbf{Limitations} While our approach performs well on concrete concepts in ConceptBench, it faces challenges in identifying and forgetting abstract concepts. Additionally, successful forgetting may require manual interventions, such as concept-specific hyperparameter tuning.
{\small
\bibliographystyle{ieee_fullname}
\bibliography{egbib}
}

\end{document}